\documentclass{bmvc2k}

\title{Importance of Self-Consistency in Active Learning for Semantic Segmentation}

\addauthor{S. Alireza Golestaneh}{sgolesta@andrew.cmu.edu}{1}
\addauthor{Kris M. Kitani}{kkitani@cs.cmu.edu}{1}

\addinstitution{
The Robotics Institute\\
Carnegie Mellon University\\
Pittsburgh, USA}

\runninghead{Golestaneh, Kitani}{Importance of Self-Consistency in Active Learning}

\begin{document}

\maketitle

\vspace{-0.15in}
\begin{abstract}
We address the task of active learning in the context of semantic segmentation and show that self-consistency can be a powerful source of self-supervision to greatly improve the performance of a data-driven model with access to only   a small amount of labeled data. Self-consistency uses the simple observation that the results of semantic segmentation for a specific image should not change under transformations like horizontal flipping (\emph{i.e.},  the results should only be flipped). In other words, the output of a model should be consistent under equivariant transformations. 
The self-supervisory signal of self-consistency is particularly helpful during active learning since the model is prone to overfitting when there is only a small amount of labeled training data. 
In our proposed active learning framework, we iteratively extract small image patches that need to be labeled, by selecting image patches that have high uncertainty (high entropy) under equivariant transformations. 
We enforce pixel-wise self-consistency between the outputs of segmentation network for each image and its transformation (horizontally flipped) to utilize the rich self-supervisory information and reduce the uncertainty of the network.
In this way, we are able to find the image patches over which the current model struggles the most to classify. By iteratively training over these difficult image patches, our experiments show that our  active learning approach reaches $\sim96\%$ of the top performance of a model trained on all data, by using only $12\%$ of the total data on benchmark semantic segmentation datasets (\emph{e.g.}, CamVid and Cityscapes).
\end{abstract}

\vspace{-0.15in}
\section{Introduction}
\vspace{-0.1in}
Deep convolutional neural networks (CNNs) are hungry for large and costly labeled training samples. CNNs have been successfully applied to different areas of research in computer vision and pattern recognition, such as classification \cite{szegedy2015going,krizhevsky2012imagenet}, detection \cite{ren2015faster}, and semantic segmentation \cite{zhao2017pyramid,ronneberger2015u,long2015fully,badrinarayanan2017segnet,chen2017deeplab,chen2017rethinking,chen2018encoder}. Semantic segmentation is one of the core tasks in visual recognition and has recently witnesses a lot of progress.

Despite the remarkable performance of existing semantic segmentation models, they have a significant drawback; they need a tremendous amount of labeled data to be able to learn their large number of parameters. A big challenge in semantic segmentation is obtaining labeled data. Active learning and weakly supervised learning \cite{khoreva2017simple,ibrahim2018weakly,dai2015boxsup} are two approaches to  mitigate the data demands of semantic segmentation.
To circumvent the burden of labeling large amounts of data, Active Learning (AL) proposed a compelling solution by iteratively selecting the \textit{most informative} samples to  label. Active learning can significantly reduce the cost of human labeling.
The goal is to focus human annotation effort towards image regions that reduce the uncertainty of the model thus improving model performance.

In this paper, we focus on the uncertainty pool-based active learning \cite{lewis1994sequential,settles2009active}, which is the most practical and commonly used approach among the existing active learning methods.
In this method, softmax probabilities from the model's outputs is used to compute entropy as an uncertainty criterion for the future manual annotation \cite{settles2009active,gal2017deep,haussmann2019deep}.  
One shortcoming of this approach is that if the model begins to overfit to the small amount of training data (\emph{i.e.}, poor generalization) gathered through active learning, the data sample selection process is negatively impacted. 
An overfit model has high uncertainty for uninformative data samples and the model makes mistakes with high certainty on other data samples. Labeling such data samples during the active learning process will hinder the model from reaching best performance. A model with poor generalization also increases the sampling bias problem \cite{beygelzimer2009importance,bach2007active,sugiyama2006active} in active learning, which decreases the quality of the selected data samples.

We tackle the problem of active learning for semantic segmentation, aiming to improve the robustness and generalization of the selected samples and model.
Our paper makes the following contributions:
i) We propose to use an equivariant transformation of the input image,  in addition to use the input image,  to improve the computed uncertainty. 
Specifically, in the active learning process, we propose to compute the uncertainty  from 
the input image and its equivariant transformed version (in this paper we use horizontal flipping as the equivariant transformation).
ii) We further propose to enforce pixel-wise self-consistency between the outputs of segmentation network for each image and its transformation to utilize the rich self-supervisory information   and reduce the uncertainty of the network.
In other words, given an input image and its  equivariant transformation we enforce the network to output similar pixel-wise predictions.
iii) We demonstrate the effectiveness of our proposed method by evaluating it on two public datasets (CamVid and Cityscapes).
Our approach achieves state-of-the-art results as it
is able to maintain better generalization and robustness over  uncertainty estimation and model generalization compared to other methods.

The rest of the paper is organized as follows. 
Some of the related works are introduced in
Section \ref{Sec2} and the proposed method is explained in Section \ref{Sec3}. Section \ref{Sec4} shows experimental
results and finally, Section \ref{Sec5} concludes the paper.



\vspace{-0.15in}
\section{Related Work}
\vspace{-0.1in}

\label{Sec2}
\textbf{Semantic segmentation} assigns a class label to each image pixel. Fully Convolutional Networks (FCNs) \cite{long2015fully} have become indispensable models for semantic segmentation. In the past few years, the accuracy of semantic segmentation has been improved rapidly \cite{badrinarayanan2017segnet,
ronneberger2015u,lin2017refinenet,
chen2017deeplab,zhao2017pyramid,
chen2018encoder} thanks to deeply learned features \cite{simonyan2014very,krizhevsky2012imagenet,he2016deep} and large-scale annotations 
\cite{lin2014microsoft,cordts2016cityscapes}.
Semantic segmentation is   critical for a variety of applications such as autonomous driving \cite{cordts2016cityscapes,ess2009segmentation,geiger2012we},   robot manipulation \cite{schwarz2018rgb}, visual question answering \cite{yu2019multi}, and medical image analysis \cite{ronneberger2015u,kuo2018cost,yang2017suggestive}. 
Two well-known challenges in semantic segmentation research are 1) acquiring dense pixel-wise labels, which is expensive and time-consuming, and 2) achieving good generalization and performance when a dataset has highly imbalanced classes.

\textbf{Active learning} tackles the problem of finding the most crucial data in a set of unlabeled samples such that the model gains the most if it is given the label of that sample. 
It has been successfully used as a method for reducing labeling costs.
A large number of active learning techniques have been developed in the literature \cite{settles2009active}.
There are mainly three  query scenarios for active learning: 
(i) membership query synthesis,
(ii) stream-based selective sampling, 
and (iii) pool-based active learning. 
In this work, we focus on pool-based active learning, which is   the most practical and common query scenario among different computer vision problems \cite{mahapatra2018efficient,siddiqui2019viewal,sinha2019variational,
yoo2019learning,geifman2019deep,pardo2019baod,heilbron2018annotate,kao2018localization,
hu2018active}.
In pool-based active learning, there exists a large unlabeled pool of data, and usually an initial small labeled set of data.
At each iteration of the process, the model is trained on the labeled set, then an acquisition function chooses one or several samples (regions in this work) from the unlabeled set to be labeled by an external oracle and added to the labeled set. 
This process is repeated until a specific budget of labeled data is exhausted or until a certain model performance is reached.

Sampling strategies in pool-based approaches have been built upon different methods, which are surveyed in \cite{settles2009active}.
Informativeness and representativeness are two important criteria when judging the importance of each sample in the unlabeled dataset. 
Informativeness measures the ability of a sample to reduce the generalization error of the adopted model, and ensure less uncertainty of the model in the next iteration, while representativeness refers to whether a sample can represent the underlying structure of the unlabeled dataset \cite{huang2010active}. 

Informative samples are usually identified by using the uncertainty of the model for each sample.
Selecting the \textit{true} informative and representative sample highly correlated with the generalizability of the model.
If the model suffers from high variance and poor generalization, it  will not identify the \textit{most informative} sample for labeling.
This   causes requesting more data to be labeled to reach the desired performance.
Different methods such as importance sampling \cite{ganti2012upal},  ensembling \cite{freund1997selective,lakshminarayanan2017simple}, Monte Carlo (MC) dropout \cite{gal2015dropout} proposed to address the aforementioned challenges.

Here we approach this problem from a different angle, our approach is not against any of the proposed methods and can be used along with them.
During the active selection, we propose to compute the uncertainty  from 
the input image and its equivariant transformed version (in this paper we use horizontal flipping as the equivariant transformation).
In this way, the computed uncertainty takes account of the model generalizability for both input and its transformation.
Therefore, the most uncertain sample will identify with more confidence.
Furthermore, 
during the model's updating stage, we propose to use pixel-wise self-consistency, in addition to cross-entropy (CE) loss,  to reduce the uncertainty of the outputs of the segmentation network between each sample and its transformation.

\textbf{Active learning for semantic segmentation}
While a variety of active learning approaches proposed for  image classification, the active learning for semantic segmentation has been less explored \cite{vijayanarasimhan2009s,vezhnevets2012active,
konyushkova2015introducing,
dutt2016active,
gorriz2017cost,yang2017suggestive,
mackowiak2018cereals,mahapatra2018efficient,
siddiqui2019viewal,kasarla2019region,casanova2020reinforced}, perhaps due to its intrinsic computational difficulties as we have to deal with much larger output spaces. 
In contrast to active learning for classification, in active learning for semantic segmentation, each image has pixel-wise predictions.
Also, the labeling cost for each image or region could be different.

Active learning for semantic segmentation with deep neural networks has been specifically investigated in \cite{gorriz2017cost,yang2017suggestive,
mackowiak2018cereals,
siddiqui2019viewal,kasarla2019region,casanova2020reinforced}.
In \cite{yang2017suggestive,gorriz2017cost}, the authors utilized MC dropout \cite{gal2015dropout}   to model the pixel-wise uncertainty and  focused on foreground-background segmentation of medical images.
Siddiqui \textit{et al.} \cite{siddiqui2019viewal} used multiple views for each image and proposed a method to estimate model's uncertainty based on the inconsistency of predictions across different views. 
In \cite{mackowiak2018cereals,kasarla2019region}, the authors designed hand-crafted methods and focused on cost-effective approaches,
where the cost of labeling for regions of an image is
not considered equal. However, labeling cost information is not always available, which 
restricts their applicability.
Casanova \textit{et al.} \cite{casanova2020reinforced} proposed an active learning strategy
for semantic segmentation based on deep reinforcement learning, where an agent
learns a policy to select a subset of image regions to label. 
Existing approaches depend entirely on the labeled regions, which had the most uncertainty during the active selection process,  to update and retrain the network.

\begin{figure}[t]
\centering
  \includegraphics [scale=.54]{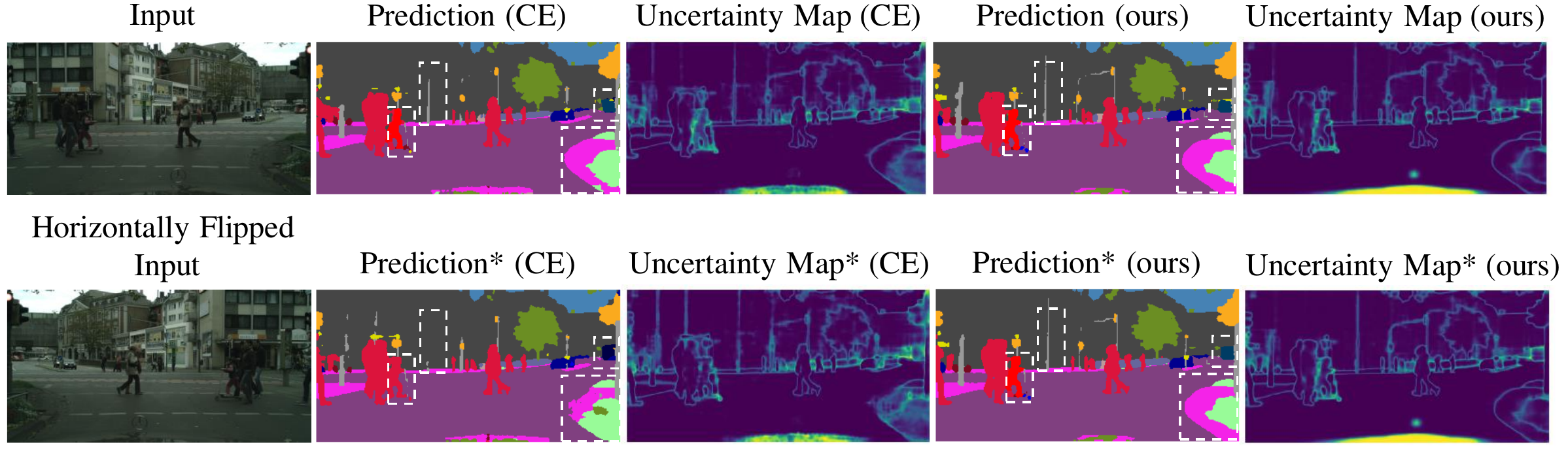}
	\caption{ Inconsistencies between segmentations.
	The input image in the second row is horizontally flipped version of the image from the first row.
	While horizontal flipping is used as an augmentation method during the training, the second column shows that the model is still suffering from poor generalization. 
	CE denotes that model only used cross-entropy loss during training.
	The fourth column shows the results while the model used our proposed self-consistency loss in addition to CE during the training. 
		* means that the image is flipped horizontally for demonstration purposes.
	For all uncertainty maps, blue indicates low values, and yellow indicates high values.
		}
	\label{SF1}
 \end{figure}

\textbf{Poor generalization of CNNs} Deep models with millions of parameters can suffer from poor generalization and high variance (specially in a low regime data situation), which   further affects the active selection process.
Overfitting and poor generalization in deep neural networks are well-known problems and an active area of research. 
Different approaches such as dropout \cite{srivastava2014dropout}, ensembling \cite{zhou2012ensemble}, and augmentation   proposed to increase the generalizability in CNNs. 
In semantic segmentation, due to large models, making predictions using a whole ensemble of models is cumbersome and is too computationally expensive to allow deployment to a large number of users.
Data augmentation has been successful in increasing the generalizability of CNNs. However, as shown in Fig. \ref{SF1}, the model can still suffer from poor generalizations for a simple transformation. 
It is shown in \cite{worrall2017harmonic,cohen2016group}  that although using different augmentation methods improve the generalization of CNNs, neural networks are still sensitive to equivalent perturbations in data. 
 Semantic segmentation models are not excepted from this issue.
For instance, as we show in Fig. \ref{SF1},  a semantic segmentation model that used image flipping as an augmentation during training still fails to have a robust prediction for an image and its flipped version.
This kind of high variance can affect the active learning process directly and causes to select the wrong sample as the most uncertain/informative one.

\textbf{Self-Supervised Learning} is a general learning framework
that relies on pretext (surrogate) tasks that can be formulated using unsupervised data.
A number of self-supervised methods have been proposed, each
exploring a different pretext task \cite{doersch2015unsupervised,gidaris2018unsupervised,hendrycks2019using,zhai2019s4l,jing2020self}. 
A pretext task is designed in a way that solving it requires learning of a useful image representation.

Inspired by the aforementioned observation and sensitivity of semantic segmentation models for equivalent transformations,  we propose  to utilize the model's uncertainty for the  input image and its equivariant transformation during the active sample selection process.
We also propose to enforce  pixel-wise self-consistency via the   self-supervisory signal between each image and its equivariant transformation to increase the robustness and generalization of the model. 
To the best of our knowledge, all current approaches for active learning in semantic segmentation
rely only on the labeled areas for updating the model.
In our proposed method, we add a new regularization that penalizes the differences between the outputs of the model for the input image and its equivariant transformation in a self-supervised manner.
Specifically, during the active selection process, we use both input and its equivariant transformation\footnote{In this paper we use horizontal flipping as the equvariant transformation.} 
to compute the uncertainty.
During the model updating stage, in addition to the cross-entropy loss, we enforce the predictions for the input sample and its transformation  to be consistent.
Our proposed approach benefits from both labeled and unlabeled regions and reduces the model's variance, which leads to a more robust sample selection.
In other words, during the active learning process, not only we use the labeled regions for updating the model, but also we use the whole image (both labeled and unlabeled regions) to enforce the  self-consistency between the image and its transformation.
Finally, we show that our proposed pixel-wise self-consistency can also be used in the final model retraining stage, after the active learning process is finished, to improve the semantic segmentation performance.
Our model is not against any of the existing active learning methods and can be used as a complementary along any of them.

\begin{figure}[t]
\centering
  \includegraphics [scale=.59]{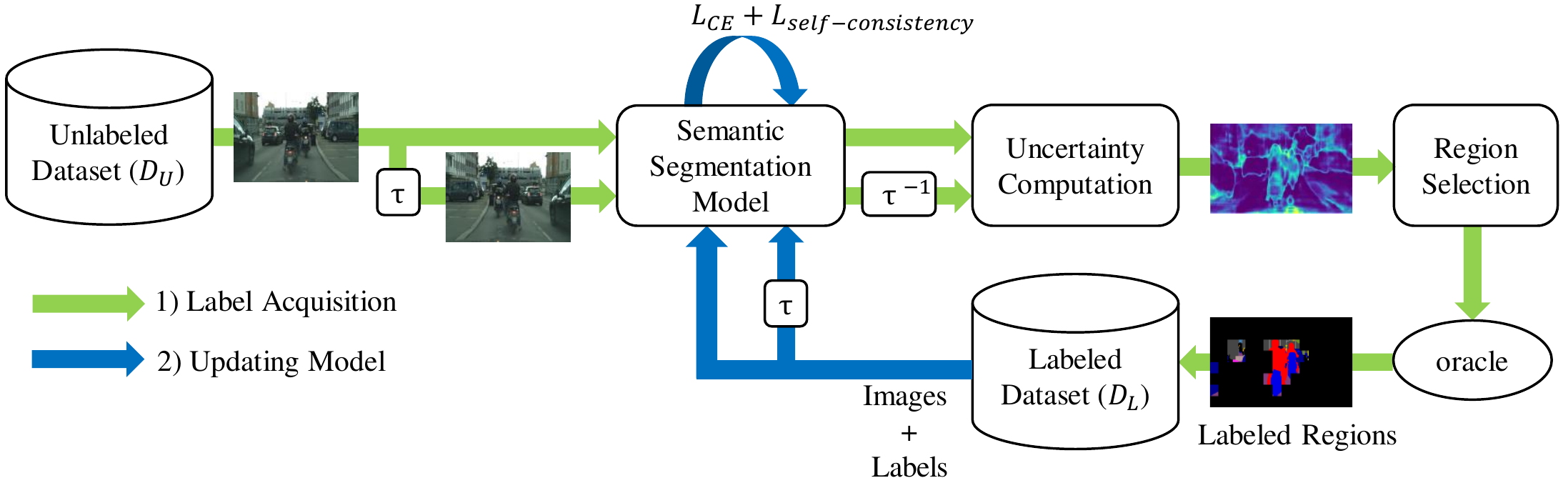}
	\caption{An overview of our proposed uncertainty pool-based active learning method.
First, at each iteration of active selection, the semantic segmentation network  computes the uncertainty for each image in $\mathcal{D}_U$ and its transformation. Then, we select K regions with the most  uncertainty and send them to an oracle for labeling. The labeled regions will be added to $\mathcal{D}_L$ and removed from $\mathcal{D}_U$.
Second,  the model   gets updated via the labeled regions in $\mathcal{D}_L$.
This is repeated until the
labeling budget is exhausted, or until a certain model performance is achieved.}
	\label{SF2}
 \end{figure}

\vspace{-0.1in}
\section{Method}
\vspace{-0.1in}
\label{Sec3}



We describe our active learning method for incorporating self-consistency over equivariant transformations of the input, to achieve high performance using only a small amount of labeled data.
Formally, let $\mathcal{D}_U$ and $\mathcal{D}_L$ denote the unlabeled and labeled datasets, respectively. Our goal is to achieve the highest performance possible through iteratively adding new labeled data until we reach a certain budget (\emph{i.e.}, maximum number of labeled data) or until a desired level of performance is achieved. Let $f_\phi$ represent the segmentation network with learnable parameters $\phi$. We select small image patches to label from unlabeled image regions such that they will maximize the performance of the segmentation model. The initial $\mathcal{D}_L$ can contain a small amount of labeled data or it can be empty. 

Fig. \ref{SF2} shows an overview of our proposed method. At each iteration $t$, the goal of active learning is to select an unlabeled image patch to be labeled and add it to the training dataset $\mathcal{D}_L$. To identify which unlabeled image patch to label, we use uncertainty sampling. Uncertainty sampling selects image regions with high uncertainty according to our current model. Regions with high uncertainty correspond to image regions that are difficult for the semantic segmentation network to classify.  Let $\mathbf{x}_i$ denote the $i^{th}$ image in $\mathcal{D}_U$. At the $t^{th}$ iteration we have two outputs cooresponding to the original image and a transformed version of the image (\emph{i.e.}, flipped image):
\vspace{-0.1in}
\begin{equation}
\label{E1}
I_{i,0}^{(u,v)}(c)=f_{t}(\mathbf{x}_{i};\phi)
\text{~and~ }
I_{i,1}^{(u,v)}(c)=\tau^{-1}(f_{t}(\tau(\mathbf{x}_{i});\phi)),
\end{equation}
\vspace{-0.1in}
\begin{equation}
\label{E2}
P_{i,j}^{(u,v)}(c)=\text{softmax}(I_{i,j}^{(u,v)}(c)),
\end{equation}
\vspace{-0.1in}
\begin{equation}
\label{E3}
H_{i}^{(u,v)}=-\sum_{j}\sum_{c}P_{i,j}^{(u,v)}(c).\log(P_{i,j}^{(u,v)}(c)),
\end{equation}
where $I_{i,0}^{(u,v)}(c)$ denotes the output logits  
of pixel $(u, v)$ belonging to class $c$ obtained from the network $f_\phi$ for an input image $\mathbf{x}_i$. $\tau$ and  $\tau^{-1}$  denote an equivariant transformation and its inverse, respectively, where $\tau^{-1}(\tau(\mathbf{x}_i))=\mathbf{x}_i$. In our experiments, we use horizontal flipping as the equivariant transformation. $P_{i,j}^{(u,v)}(c)$ is the softmax probability of pixel $(u, v)$ belonging to class $c$  obtained from $I_{i,j}^{(u,v)}(c)$, where $j \in \{0,1\}$. Interestingly, for many semantic segmentation networks, flipping the image will results in slightly different results such that $I_{i,0}^{(u,v)}(c) \neq I_{i,1}^{(u,v)}(c)$. In Eq. \ref{E3}, $H_{i}^{(u,v)}$ denotes the pixel-wise entropy (estimated uncertainty) for $i^{th}$ image, where it is computed by aggregating the posterior uncertainties over network predictions for image $\mathbf{x}_i$ and its transformation. $H_{i}^{(u,v)}$ captures the uncertainty of the network not only for input $\mathbf{x}_i$, but also for its transformation.

To exploit the structure and locality in semantic segmentation, we argue for selecting regions to be labeled instead of whole images. 
In particular, we opt for using fixed sized rectangular regions.
Given $H_i$ as the pixel-wise entropy for image $i$, we compute the next region to label by dividing $H_i$ into $M$ fixed-size rectangular regions denoted by $H_i^m$ where $m\in M$.
We then look for the region among all the unlabeled data that has the highest entropy:
\begin{equation}
\label{E4}
(s,n)=\arg\max_{(i,m)}H_{i}^{m},
\end{equation}
where $H^n_s$ has the maximum uncertainty among the unlabeled data.
The selected region  will not be considered for further
selection. 
Eq. \ref{E4} will be repeated K times, where K is the number of samples that are set to be labeled at each iteration by an oracle.

In our experiments, during the label acquisition process, instead of using a real annotator, we simulate
annotation by using the ground truth annotation of the region as the annotation from the oracle. 
Images which contained labeled
regions are then added to the labeled dataset.
If an image is fully labeled, it will be removed
from the unlabeled dataset. 
The labeled dataset, therefore,
is comprised of a lot of images, each with a subset of their
regions labeled. 
The unlabeled regions of these images are marked with the ignore label.
For each image $x_l$ where $l\in |\mathcal{D}_L|$ we have the ground truth $y_l$.
After each label acquisition iteration, the network will get updated via the labeled data.
In all of the existing active learning works for semantic segmentation, the network gets updated via the labeled data to minimize the cross-entropy loss.
However, for each image, there are unlabeled regions with rich information that can be exploited to reduce the uncertainty of the network.
In this work, we propose to use both labeled and unlabeled regions to increase the self-consistency of the network.
At each iteration, after each acquisition step, the semantic segmentation network will
retrain as following:

\vspace{-0.2in}
\begin{subequations}
\label{E5}
\begin{align}
\min_{\phi}L_{\text{total}}\text{       }\;\text{where}\;\;L_{\text{total}}=L_{\text{CE}}+  L_{\text{self-consistency}}, \\
L_{\text{CE}}=-\sum_{j}\sum_{h,w,c}y_{l}.\log(P_{i,j}), \\
L_{\text{self-consistency}}=\frac{2}{W\times H\times C}\parallel I_{i,0}^{(u,v)}(c)-I_{i,1}^{(u,v)}(c)\parallel_{2}^{2},
\end{align} 
\end{subequations}
where $L_{CE}$  is the standard supervised pixel-wise cross-entropy loss, 
where it is evaluated only for the regions that are labeled.
Eq. (\ref{E5}c) is the self-consistency loss, which is used to utilize the information between the outputs of the model for input image and its corresponding transformation.
In Eq. (\ref{E5}c), $W$, $H$, $C$ denote width, height, and depth of the logits, respectively.
Enforcing the self-consistency between the outputs of model for the input and its transformation helps the network generalize better and have less variance.

After meeting the budget amount, the active selection process concludes with the retraining of the network with the final dataset $\mathcal{D}_L$, we refer that as the final retraining stage.  

\vspace{-0.1in}
\section{Results}
\vspace{-0.1in}
\label{Sec4}
\subsection{Datasets}
We evaluate our method on two semantic segmentation datasets: CamVid  and Cityscapes.\\ 
\textbf{CamVid} dataset \cite{brostow2009semantic}  composed of street scene view images, where each images has $360 \times 480$ resolution and contains 11 semantic  classes. 
CamVid has 367, 101 and 233 images for train, validation, and test
set, respectively. 
\textbf{Cityscapes} dataset \cite{cordts2016cityscapes} is a large dataset that focuses on
semantic understanding of urban street scenes. The dataset
contains $2048 \times 1024$  images with fine annotations across different cities,
different seasons, and varying scene layout and background.
The dataset is annotated with 30 categories, of which 19
categories are included for training and evaluation (others
are ignored). 
The training, validation, and test set contains
2975, 500, and 1525 fine images, respectively.

\subsection{Implementation Details}
In our experiments,  for all the methods, we use the same segmentation network  (ResNet50 \cite{long2015fully} + FCN \cite{he2016deep}).\footnote{Since we do not change the network architecture, any other segmentation network can be used.}  
To analyze the true potential of each active learning method, for all the experiments, for the active sample selection process, the segmentation network is initialized with random weights.
After the  active learning process is finished, we retrain the networks (i.e. the final retraining stage) with the  final dataset $\mathcal{D}_L$  while the ResNet50 backbone is initialized with weights
from a model that was pre-trained on the ILSVRC
1000-class classification task \cite{russakovsky2015imagenet}.
For all the experiments, we
use   random horizontal flipping for the
data augmentation.
For the final retraining stage, we use 60 epochs. We set the batch size to five and four for CamVid and Cityscapes, respectively.
For optimization, we use the adaptive moment estimation optimizer
(ADAM) \cite{kingma2014adam} with $\beta_1 = 0.9$, $\beta_2 = 0.999$, we set the initial learning
rate and weight decay to 5e-4 and 2e-4, respectively.
Learning rate decays is set to 0.1, and it applied after every 20 epochs. 
For all the experiments, we report the average  of the five different runs. 

For both datasets, we use their training sets as $\mathcal{D}_U$.
For CamVid we evaluate the performance on the test set and report the mIoU.
We use $360\times 480$  images as an input of the network; each image is divided into 25 non-overlapping regions of dimension $72\times 96$.
We chose
$K = 1$ region per step. 
For Cityscapes  
we
report the mIoU on the validation set since the test set
is not available publicly.
We resize the images to $512\times 1024$  because of the computational complexity.
Each image is divided into 32 non-overlapping regions of dimension $128\times 128$.
We chose
$K = 5$ regions per step.

\vspace{-0.1in}
\subsection{Experiments}
\begin{figure}[t]
\centering
  \includegraphics [scale=.39]{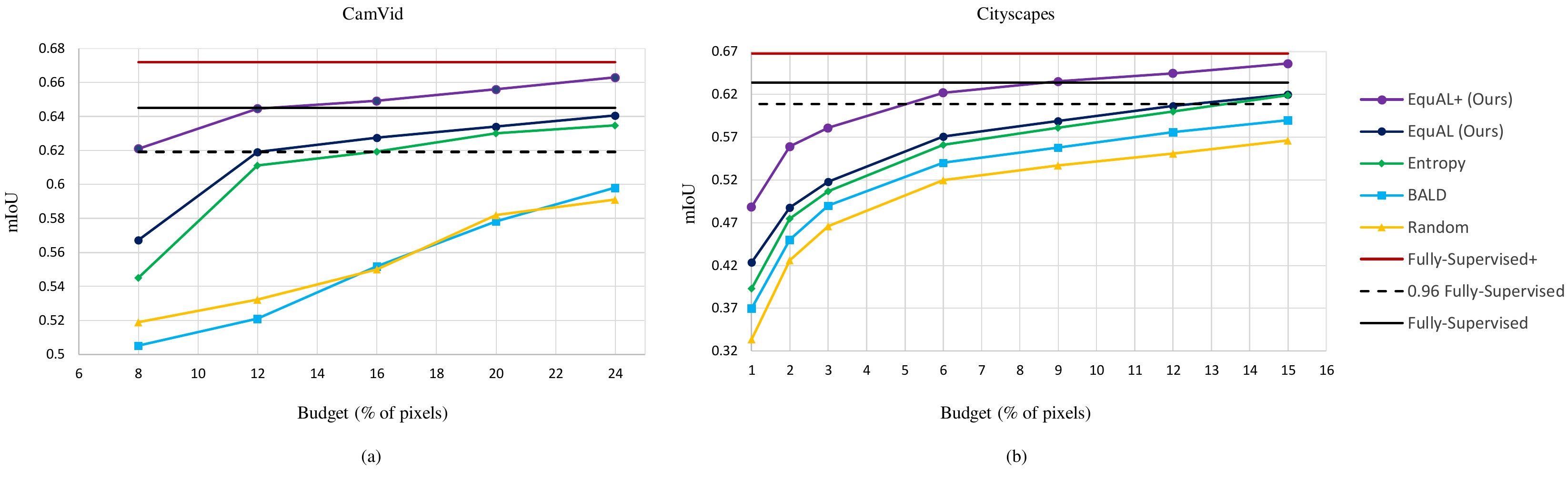}
	\caption{ Active learning performance for our proposed method and other baselines. 
	(a) and (b) demonstrate the mIoU results for CamVid and Cityscapes datasets, respectively.
	The horizontal solid lines  at the top
represent the performance with the entire training set labeled. The dashed line  represents $96\%$ of the maximum model performance (Fully-Supervised).
 Budget denotes   the
percentage of unlabeled data labeled during the active learning process. }
	\label{SF3}
 \end{figure}

Fig. \ref{SF3} demonstrates the mIoU results achieved by the segmentation
model against the percentage of datasets labeled used to train
the model.  
We evaluate our proposed method for different budget sizes against different baselines on two datasets.
During the active selection, we use the same settings for all the methods.
After collecting the labeled data,   models are retrained in the same environment and hyperparameters.

 Fig. \ref{SF3} summarizes the results of our experiments.
 In our experiments, at the final retraining stage, Fully-Supervised, Random, Entropy, and EquAL baselines are retrained via  CE loss.
 Fully-Supervised+ and EquAL+ baselines are  retrained via  CE loss as well as the proposed pixel-wise self-conssitency loss (Eq. \ref{E5}). 

Fig. \ref{SF3}  demonstrates that our proposed methods, EquAL and EquAL+,  outperforms the  other baselines for the same percentage of
labeled data.
We observe that for the low budget size, in both of the datasets, all the baselines suffer from the cold start problem \cite{houlsby2014cold,shao2019learning}.
In contrast, our method achieves significantly better performance in the low regime data, which we believe is due to the better generalization.
Particularly, in CamVid, for $8\%$ budget  we improved the entropy baseline by $2\%$ (EquAL) and $7\%$ (EquAL+).
 In Cityscapes, for $1\%$ budget  we improved the entropy baseline by $3\%$ (EquAL) and $9\%$ (EquAL+).
 As we move to  larger budgets the networks can achieve better generalization.
EquAL achieves $96\%$ of the maximum model performance with
$12\%$   labeled data in both datasets, while the runner-up
method does so with $16\%$ and $13\%$ data on the respective
datasets, which corresponds to 14 and 29 additional images, respectively.
Furthermore, we show that by using the self-consistency loss in addition to the CE loss we can consistency achieves better performance.
To measure the diversity of the labeled data, for $12\%$ budget in the Cityscapes,  we further compute the entropy of the class distribution for the data in the final labeled set;
our model achieves higher entropy (2.376) comparing to the entropy baseline (2.335),  which means more diversity in the selected samples.

Table \ref{TB1} provides per-class IoU comparison between our model and various baselines, where the budget is fixed to $12\%$. 
All the models except EquAL+ uses CE loss during the final retraining stage.
Table \ref{TB1} shows that our proposed method, EquAL, outperforms other methods on 5 classes as well as mIoU and achieves competitive results  for the other classes.
EquAL+ denotes our model when our self-consistency loss is used in addition to CE loss during the final retraining stage and leveraged both labeled and unlabeled regions.
In EquAL+, we regularize the dark knowledge of the model and reduces the high variance, which leads to a significantly better performance.

\begin{table}[h]
\centering
\caption{Per-class performance comparison results for Citydataset while setting the labeling budget to $12\%$. Bold   entries are
the best and second-best performers.}
\resizebox{5 in}{!} {
\begin{tabular}{cccccccccc|c}
\hline 
Methods & Road & SideWalk & Building & Wall & Fence & Pole & Traffic Light & Traffic Sign & \multicolumn{1}{c}{Vegetation} & Terrain\tabularnewline
\hline 
\hline 
Random & \textbf{96.64} & \textbf{75.51} & \textbf{88.19} & 32.51 & 34.61 & 36.09 & 38.80 & 58.25 & \multicolumn{1}{c}{89.03} & 53.48\tabularnewline
BALD \cite{gal2017deep} & 96.22 & 74.08 & 87.71 & 33.68 & 39.73 & 40.78 & 44.90 & 40.67 & \multicolumn{1}{c}{89.13} & 52.79\tabularnewline
Vote Entropy \cite{mackowiak2018cereals} & 96.22 & 73.50 & 87.08 & 34.97 & \textbf{42.32} & 42.76 & 44.91 & 61.32 & \multicolumn{1}{c}{\textbf{89.24}} & 54.18\tabularnewline
RALIS \cite{casanova2020reinforced} & 96.19 & 73.24 & 86.25 & 33.56 & 42.28 & 43.51 & \textbf{51.12} & 60.31 & \multicolumn{1}{c}{89.05} & 51.72\tabularnewline
Entropy & 96.11 & 73.72 & 87.78 & 33.76 & 40.40 & \textbf{44.50} & 44.78 & 61.32 & \multicolumn{1}{c}{89.22} & 53.77\tabularnewline
EquAL (Ours) & 96.05 & 73.05 & 87.66 & \textbf{40.94} & 42.31 & 43.79 & 46.22 & \textbf{61.58} & \multicolumn{1}{c}{89.14} & \textbf{56.15}\tabularnewline
EquAL+ (Ours) & \textbf{97.01} & \textbf{77.90} & \textbf{89.06} & \textbf{41.74} & \textbf{44.93} & \textbf{46.07} & \textbf{51.78} & \textbf{65.44} & \multicolumn{1}{c}{\textbf{89.88}} & \textbf{58.33}\tabularnewline
\hline 
\multirow{2}{*}{} & \multirow{2}{*}{Sky} & \multirow{2}{*}{Person} & \multirow{2}{*}{Rider} & \multirow{2}{*}{Car} & \multirow{2}{*}{Truck} & \multirow{2}{*}{Bus} & \multirow{2}{*}{Train} & \multirow{2}{*}{Motorcycle} & \multirow{2}{*}{Bicycle} & \multirow{2}{*}{mIoU}\tabularnewline
 &  &  &  &  &  &  &  &  &  & \tabularnewline
\hline 
\hline 
Random & 90.18 & 52.67 & 32.51 & 88.47 & 24.62 & 41.93 & 31.26 & 29.11 & 60.15 & 55.47\tabularnewline
BALD \cite{gal2017deep} & 89.81 & 65.73 & 39.52 & \textbf{89.54} & 34.65 & 55.59 & 36.14 & 31.27 & 63.33 & 58.17\tabularnewline
Vote Entropy \cite{mackowiak2018cereals} & 89.94 & \textbf{69.31} & 40.59 & 89.14 & 37.09 & 54.92 & \textbf{36.57} & 32.16 & 65.73 & 60.01\tabularnewline
RALIS \cite{casanova2020reinforced} & 89.32 & 69.23 & \textbf{42.14} & 89.51 & \textbf{39.50} & 57.22 & 35.51 & \textbf{31.42} & 64.12 & 60.27\tabularnewline
Entropy & 90.53 & 68.81 & 41.09 & 89.52 & 37.87 & 54.83 & 36.15 & 31.03 & \textbf{65.51} & 60.03\tabularnewline
EquAL (Ours) & \textbf{90.58} & 68.10 & 40.75 & 89.32 & 37.39 & \textbf{58.63} & 35.03 & 31.27 & 64.49 & \textbf{60.65}\tabularnewline
EquAL+ (Ours) & \textbf{92.05 } & \textbf{72.30 } & \textbf{47.74 } & \textbf{91.06} & \textbf{48.53 } & \textbf{63.75} & \textbf{36.98} & \textbf{42.19 } & \textbf{68.11 } & \textbf{64.46}\tabularnewline
\hline 
\end{tabular}
}
\label{TB1}
\end{table}

\begin{figure}[h]
\centering
  \includegraphics [scale=.37]{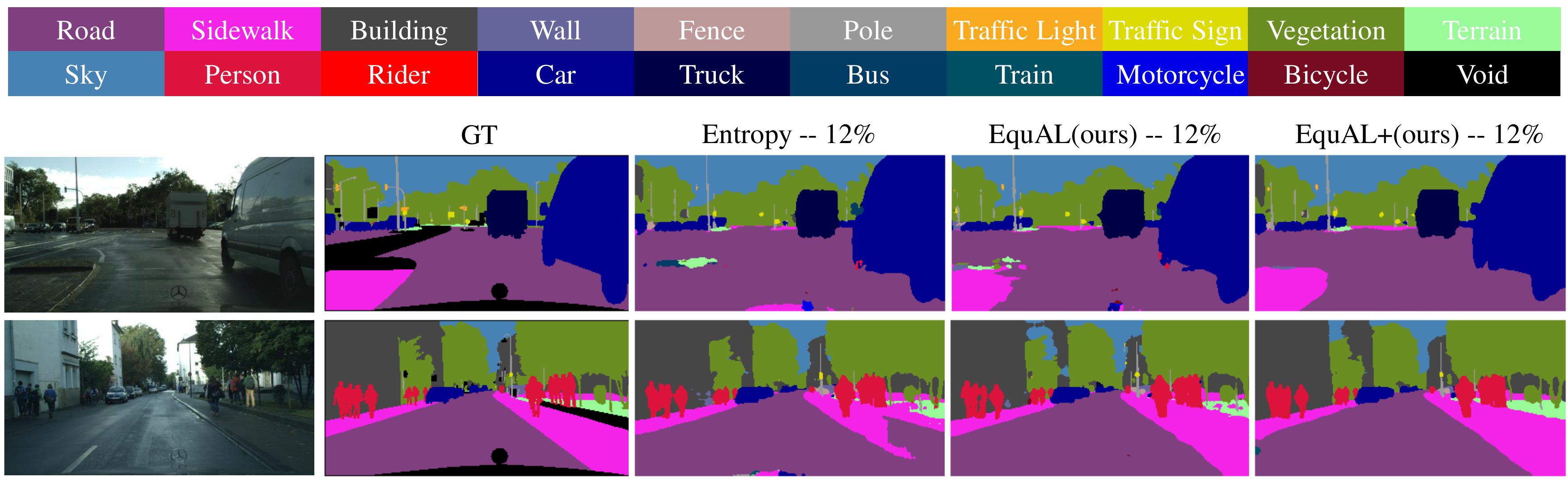}
	\caption{Qualitative results on the Cityscpaes dataset. 
	All the models are retrained after they use active learning  with budget limit of $12\%$.
The third column represent the entropy baseline, and last two columns represent our proposed models.
	}
	\label{SF4} 
 \end{figure}

Fig. \ref{SF4} provides qualitative comparison between our model and entropy baseline for the same budget ($12\%$).
We observe that the entropy baseline (third column) fails to segment sidewalk and terrain comparing to our proposed model (forth column), EquAL.
Moreover, we  show in the fifth column (EquAL+) that we can even  improve our results further by adding the pixel-wise self-consistency loss.

To study the effects of different equivariant transformations, in Table \ref{TB3},  we study  effects of employing five different equivariant transformations during the active learning process while using CamVid. It is not surprising to see that translation and horizontal flipping are the only two that improves the performance.
We believe that this behavior is because transformations such as rotation or vertical flipping create images with conditions that do not exist in the dataset.

\begin{table}[h]
\centering
\caption{Ablation study for different equivariant transformations during the active learning process.}
\resizebox{3. in}{!} {
\begin{tabular}{ccccc}
\hline 
\multicolumn{5}{c}{CamVid}\tabularnewline
\hline 
\hline 
\multirow{2}{*}{Models} & \multicolumn{4}{c}{Budget}\tabularnewline
\cline{2-5} \cline{3-5} \cline{4-5} \cline{5-5} 
 & 8\% & 12\% & 16\% & 20\%\tabularnewline
\hline 
Entropy (Baseline) & 0.5451 & 0.6111 & 0.6192 & 0.6311\tabularnewline
EquAL (Rotation 90 degree) & 0.5158 & 0.5855 & 0.5879 & 0.6036\tabularnewline
EquAL (Rotation 180 degree) & 0.5278 & 0.5982 & 0.6005 & 0.6072\tabularnewline
EquAL (Random Translation 16 pixels) & 0.5472 & 0.6121 & 0.6193 & 0.6294\tabularnewline
EquAL (Vertical Flipping) & 0.5266 & 0.5925 & 0.5987 & 0.6027\tabularnewline
EquAL (Horizental Flipping) & 0.5673 & 0.6182 & 0.6274 & 0.6341\tabularnewline
\hline 
 &  &  &  & \tabularnewline
\end{tabular}
}
\label{TB3}
\end{table}

\vspace{-0.3in}
\section{Conclusion}
\vspace{-0.1in}
\label{Sec5}
In this paper, we proposed an effective method for active learning for semantic segmentation.
We propose to use an equivariant transformation of the input image, in addition to use the input image, to improve the computed uncertainty during the active learning process.
We also proposed a pixel-wise self-consistency loss, which takes advantage of self-supervisory information and regularizes the inconsistency of the model predictions between the input image and its transformation.
Our proposed method yields a more robust active  sample selection   as well as better segmentation performance.
Specifically, our method achieves $\sim96\%$ of maximum achievable network
performance using only $12\%$   labeled data
on CamVid and Cityscapes datasets. Finally, by  using our proposed pixel-wise self-consistency we improve the segmentation performance by $2$-$9\%$.

 As
future work, we highlight the possibility of designing an optimal stopping method  that could help
accelerate the active sample selection process, in this way the active learning algorithms do not require to iterate  completely through the unlabeled set to select the most uncertain samples.
 Furthermore, if the  community provides  ground-truth information about human annotation costs for more datasets, it will be an important direction to develop  cost-estimate  and cost-effective methods to reduce the cost of labeling for differemt regions even further.
 
\vspace{-0.1in}
\section{Acknowledgement}
\vspace{-0.1in}
This project was sponsored by Caterpillar Inc.
\bibliography{egbib}

\begin{thebibliography}{67}
\providecommand{\natexlab}[1]{#1}
\providecommand{\url}[1]{\texttt{#1}}
\expandafter\ifx\csname urlstyle\endcsname\relax
  \providecommand{\doi}[1]{doi: #1}\else
  \providecommand{\doi}{doi: \begingroup \urlstyle{rm}\Url}\fi

\bibitem[Bach(2007)]{bach2007active}
Francis~R Bach.
\newblock Active learning for misspecified generalized linear models.
\newblock In \emph{Advances in neural information processing systems}, pages
  65--72, 2007.

\bibitem[Badrinarayanan et~al.(2017)Badrinarayanan, Kendall, and
  Cipolla]{badrinarayanan2017segnet}
Vijay Badrinarayanan, Alex Kendall, and Roberto Cipolla.
\newblock Segnet: A deep convolutional encoder-decoder architecture for image
  segmentation.
\newblock \emph{IEEE transactions on pattern analysis and machine
  intelligence}, 39\penalty0 (12):\penalty0 2481--2495, 2017.

\bibitem[Beygelzimer et~al.(2009)Beygelzimer, Dasgupta, and
  Langford]{beygelzimer2009importance}
Alina Beygelzimer, Sanjoy Dasgupta, and John Langford.
\newblock Importance weighted active learning.
\newblock In \emph{Proceedings of the 26th annual international conference on
  machine learning}, pages 49--56, 2009.

\bibitem[Brostow et~al.(2009)Brostow, Fauqueur, and
  Cipolla]{brostow2009semantic}
Gabriel~J Brostow, Julien Fauqueur, and Roberto Cipolla.
\newblock Semantic object classes in video: A high-definition ground truth
  database.
\newblock \emph{Pattern Recognition Letters}, 30\penalty0 (2):\penalty0 88--97,
  2009.

\bibitem[Casanova et~al.(2020)Casanova, Pinheiro, Rostamzadeh, and
  Pal]{casanova2020reinforced}
Arantxa Casanova, Pedro~O Pinheiro, Negar Rostamzadeh, and Christopher~J Pal.
\newblock Reinforced active learning for image segmentation.
\newblock \emph{arXiv preprint arXiv:2002.06583}, 2020.

\bibitem[Chen et~al.(2017{\natexlab{a}})Chen, Papandreou, Kokkinos, Murphy, and
  Yuille]{chen2017deeplab}
Liang-Chieh Chen, George Papandreou, Iasonas Kokkinos, Kevin Murphy, and Alan~L
  Yuille.
\newblock Deeplab: Semantic image segmentation with deep convolutional nets,
  atrous convolution, and fully connected crfs.
\newblock \emph{IEEE transactions on pattern analysis and machine
  intelligence}, 40\penalty0 (4):\penalty0 834--848, 2017{\natexlab{a}}.

\bibitem[Chen et~al.(2017{\natexlab{b}})Chen, Papandreou, Schroff, and
  Adam]{chen2017rethinking}
Liang-Chieh Chen, George Papandreou, Florian Schroff, and Hartwig Adam.
\newblock Rethinking atrous convolution for semantic image segmentation.
\newblock \emph{arXiv preprint arXiv:1706.05587}, 2017{\natexlab{b}}.

\bibitem[Chen et~al.(2018)Chen, Zhu, Papandreou, Schroff, and
  Adam]{chen2018encoder}
Liang-Chieh Chen, Yukun Zhu, George Papandreou, Florian Schroff, and Hartwig
  Adam.
\newblock Encoder-decoder with atrous separable convolution for semantic image
  segmentation.
\newblock In \emph{Proceedings of the European conference on computer vision
  (ECCV)}, pages 801--818, 2018.

\bibitem[Cohen and Welling(2016)]{cohen2016group}
Taco Cohen and Max Welling.
\newblock Group equivariant convolutional networks.
\newblock In \emph{International conference on machine learning}, pages
  2990--2999, 2016.

\bibitem[Cordts et~al.(2016)Cordts, Omran, Ramos, Rehfeld, Enzweiler, Benenson,
  Franke, Roth, and Schiele]{cordts2016cityscapes}
Marius Cordts, Mohamed Omran, Sebastian Ramos, Timo Rehfeld, Markus Enzweiler,
  Rodrigo Benenson, Uwe Franke, Stefan Roth, and Bernt Schiele.
\newblock The cityscapes dataset for semantic urban scene understanding.
\newblock In \emph{Proceedings of the IEEE conference on computer vision and
  pattern recognition}, pages 3213--3223, 2016.

\bibitem[Dai et~al.(2015)Dai, He, and Sun]{dai2015boxsup}
Jifeng Dai, Kaiming He, and Jian Sun.
\newblock Boxsup: Exploiting bounding boxes to supervise convolutional networks
  for semantic segmentation.
\newblock In \emph{Proceedings of the IEEE international conference on computer
  vision}, pages 1635--1643, 2015.

\bibitem[Doersch et~al.(2015)Doersch, Gupta, and
  Efros]{doersch2015unsupervised}
Carl Doersch, Abhinav Gupta, and Alexei~A Efros.
\newblock Unsupervised visual representation learning by context prediction.
\newblock In \emph{Proceedings of the IEEE international conference on computer
  vision}, pages 1422--1430, 2015.

\bibitem[Dutt~Jain and Grauman(2016)]{dutt2016active}
Suyog Dutt~Jain and Kristen Grauman.
\newblock Active image segmentation propagation.
\newblock In \emph{Proceedings of the IEEE Conference on Computer Vision and
  Pattern Recognition}, pages 2864--2873, 2016.

\bibitem[Ess et~al.()Ess, M{\"u}ller, Grabner, and
  Van~Gool]{ess2009segmentation}
Andreas Ess, Tobias M{\"u}ller, Helmut Grabner, and Luc~J Van~Gool.
\newblock Segmentation-based urban traffic scene understanding.
\newblock Citeseer.

\bibitem[Freund et~al.(1997)Freund, Seung, Shamir, and
  Tishby]{freund1997selective}
Yoav Freund, H~Sebastian Seung, Eli Shamir, and Naftali Tishby.
\newblock Selective sampling using the query by committee algorithm.
\newblock \emph{Machine learning}, 28\penalty0 (2-3):\penalty0 133--168, 1997.

\bibitem[Gal and Ghahramani(2016)]{gal2015dropout}
Yarin Gal and Zoubin Ghahramani.
\newblock Dropout as a bayesian approximation: Representing model uncertainty
  in deep learning.
\newblock \emph{In international conference on machine learning}, pages
  740--755, 2016.

\bibitem[Gal et~al.(2017)Gal, Islam, and Ghahramani]{gal2017deep}
Yarin Gal, Riashat Islam, and Zoubin Ghahramani.
\newblock Deep bayesian active learning with image data.
\newblock In \emph{Proceedings of the 34th International Conference on Machine
  Learning-Volume 70}, pages 1183--1192. JMLR. org, 2017.

\bibitem[Ganti and Gray(2012)]{ganti2012upal}
Ravi Ganti and Alexander Gray.
\newblock Upal: Unbiased pool based active learning.
\newblock In \emph{Artificial Intelligence and Statistics}, pages 422--431,
  2012.

\bibitem[Geifman and El-Yaniv(2019)]{geifman2019deep}
Yonatan Geifman and Ran El-Yaniv.
\newblock Deep active learning with a neural architecture search.
\newblock In \emph{Advances in Neural Information Processing Systems}, pages
  5974--5984, 2019.

\bibitem[Geiger et~al.(2012)Geiger, Lenz, and Urtasun]{geiger2012we}
Andreas Geiger, Philip Lenz, and Raquel Urtasun.
\newblock Are we ready for autonomous driving? the kitti vision benchmark
  suite.
\newblock In \emph{2012 IEEE Conference on Computer Vision and Pattern
  Recognition}, pages 3354--3361. IEEE, 2012.

\bibitem[Gidaris et~al.(2018)Gidaris, Singh, and
  Komodakis]{gidaris2018unsupervised}
Spyros Gidaris, Praveer Singh, and Nikos Komodakis.
\newblock Unsupervised representation learning by predicting image rotations.
\newblock \emph{arXiv preprint arXiv:1803.07728}, 2018.

\bibitem[Gorriz et~al.(2017)Gorriz, Carlier, Faure, and Giro-i
  Nieto]{gorriz2017cost}
Marc Gorriz, Axel Carlier, Emmanuel Faure, and Xavier Giro-i Nieto.
\newblock Cost-effective active learning for melanoma segmentation.
\newblock \emph{arXiv preprint arXiv:1711.09168}, 2017.

\bibitem[Hau{\ss}mann et~al.(2019)Hau{\ss}mann, Hamprecht, and
  Kandemir]{haussmann2019deep}
Manuel Hau{\ss}mann, Fred Hamprecht, and Melih Kandemir.
\newblock Deep active learning with adaptive acquisition.
\newblock In \emph{Proceedings of the 28th International Joint Conference on
  Artificial Intelligence}, pages 2470--2476. AAAI Press, 2019.

\bibitem[He et~al.(2016)He, Zhang, Ren, and Sun]{he2016deep}
Kaiming He, Xiangyu Zhang, Shaoqing Ren, and Jian Sun.
\newblock Deep residual learning for image recognition.
\newblock In \emph{Proceedings of the IEEE conference on computer vision and
  pattern recognition}, pages 770--778, 2016.

\bibitem[Heilbron et~al.(2018)Heilbron, Lee, Jin, and
  Ghanem]{heilbron2018annotate}
Fabian~Caba Heilbron, Joon-Young Lee, Hailin Jin, and Bernard Ghanem.
\newblock What do i annotate next? an empirical study of active learning for
  action localization.
\newblock In \emph{European Conference on Computer Vision}, pages 212--229.
  Springer, 2018.

\bibitem[Hendrycks et~al.(2019)Hendrycks, Mazeika, Kadavath, and
  Song]{hendrycks2019using}
Dan Hendrycks, Mantas Mazeika, Saurav Kadavath, and Dawn Song.
\newblock Using self-supervised learning can improve model robustness and
  uncertainty.
\newblock In \emph{Advances in Neural Information Processing Systems}, pages
  15663--15674, 2019.

\bibitem[Houlsby et~al.(2014)Houlsby, Hern{\'a}ndez-Lobato, and
  Ghahramani]{houlsby2014cold}
Neil Houlsby, Jos{\'e}~Miguel Hern{\'a}ndez-Lobato, and Zoubin Ghahramani.
\newblock Cold-start active learning with robust ordinal matrix factorization.
\newblock In \emph{International Conference on Machine Learning}, pages
  766--774, 2014.

\bibitem[Hu et~al.(2018)Hu, Lipton, Anandkumar, and Ramanan]{hu2018active}
Peiyun Hu, Zachary~C Lipton, Anima Anandkumar, and Deva Ramanan.
\newblock Active learning with partial feedback.
\newblock \emph{arXiv preprint arXiv:1802.07427}, 2018.

\bibitem[Huang et~al.(2010)Huang, Jin, and Zhou]{huang2010active}
Sheng-Jun Huang, Rong Jin, and Zhi-Hua Zhou.
\newblock Active learning by querying informative and representative examples.
\newblock In \emph{Advances in neural information processing systems}, pages
  892--900, 2010.

\bibitem[Ibrahim et~al.(2020)Ibrahim, Vahdat, Ranjbar, and
  Macready]{ibrahim2018weakly}
Mostafa~S Ibrahim, Arash Vahdat, Mani Ranjbar, and William~G Macready.
\newblock Weakly supervised semantic image segmentation with self-correcting
  networks.
\newblock In \emph{Proceedings of the IEEE conference on computer vision and
  pattern recognition}, pages 876--885, 2020.

\bibitem[Jing and Tian(2020)]{jing2020self}
Longlong Jing and Yingli Tian.
\newblock Self-supervised visual feature learning with deep neural networks: A
  survey.
\newblock \emph{IEEE Transactions on Pattern Analysis and Machine
  Intelligence}, 2020.

\bibitem[Kao et~al.(2018)Kao, Lee, Sen, and Liu]{kao2018localization}
Chieh-Chi Kao, Teng-Yok Lee, Pradeep Sen, and Ming-Yu Liu.
\newblock Localization-aware active learning for object detection.
\newblock In \emph{Asian Conference on Computer Vision}, pages 506--522.
  Springer, 2018.

\bibitem[Kasarla et~al.(2019)Kasarla, Nagendar, Hegde, Balasubramanian, and
  Jawahar]{kasarla2019region}
Tejaswi Kasarla, G~Nagendar, Guruprasad~M Hegde, V~Balasubramanian, and
  CV~Jawahar.
\newblock Region-based active learning for efficient labeling in semantic
  segmentation.
\newblock In \emph{2019 IEEE Winter Conference on Applications of Computer
  Vision (WACV)}, pages 1109--1117. IEEE, 2019.

\bibitem[Khoreva et~al.(2017)Khoreva, Benenson, Hosang, Hein, and
  Schiele]{khoreva2017simple}
Anna Khoreva, Rodrigo Benenson, Jan Hosang, Matthias Hein, and Bernt Schiele.
\newblock Simple does it: Weakly supervised instance and semantic segmentation.
\newblock In \emph{Proceedings of the IEEE conference on computer vision and
  pattern recognition}, pages 876--885, 2017.

\bibitem[Kingma and Ba(2014)]{kingma2014adam}
Diederik~P Kingma and Jimmy Ba.
\newblock Adam: A method for stochastic optimization.
\newblock \emph{arXiv preprint arXiv:1412.6980}, 2014.

\bibitem[Konyushkova et~al.(2015)Konyushkova, Sznitman, and
  Fua]{konyushkova2015introducing}
Ksenia Konyushkova, Raphael Sznitman, and Pascal Fua.
\newblock Introducing geometry in active learning for image segmentation.
\newblock In \emph{Proceedings of the IEEE International Conference on Computer
  Vision}, pages 2974--2982, 2015.

\bibitem[Krizhevsky et~al.(2012)Krizhevsky, Sutskever, and
  Hinton]{krizhevsky2012imagenet}
Alex Krizhevsky, Ilya Sutskever, and Geoffrey~E Hinton.
\newblock Imagenet classification with deep convolutional neural networks.
\newblock In \emph{Advances in neural information processing systems}, pages
  1097--1105, 2012.

\bibitem[Kuo et~al.(2018)Kuo, H{\"a}ne, Yuh, Mukherjee, and Malik]{kuo2018cost}
Weicheng Kuo, Christian H{\"a}ne, Esther Yuh, Pratik Mukherjee, and Jitendra
  Malik.
\newblock Cost-sensitive active learning for intracranial hemorrhage detection.
\newblock In \emph{International Conference on Medical Image Computing and
  Computer-Assisted Intervention}, pages 715--723. Springer, 2018.

\bibitem[Lakshminarayanan et~al.(2017)Lakshminarayanan, Pritzel, and
  Blundell]{lakshminarayanan2017simple}
Balaji Lakshminarayanan, Alexander Pritzel, and Charles Blundell.
\newblock Simple and scalable predictive uncertainty estimation using deep
  ensembles.
\newblock In \emph{Proceedings of the 31st International Conference on Neural
  Information Processing Systems}, pages 6405--6416. Curran Associates Inc.,
  2017.

\bibitem[Lewis and Gale(1994)]{lewis1994sequential}
David~D Lewis and William~A Gale.
\newblock A sequential algorithm for training text classifiers.
\newblock In \emph{Proceedings of the 17th annual international ACM SIGIR
  conference on Research and development in information retrieval}, pages
  3--12, 1994.

\bibitem[Lin et~al.(2017)Lin, Milan, Shen, and Reid]{lin2017refinenet}
Guosheng Lin, Anton Milan, Chunhua Shen, and Ian Reid.
\newblock Refinenet: Multi-path refinement networks for high-resolution
  semantic segmentation.
\newblock In \emph{Proceedings of the IEEE conference on computer vision and
  pattern recognition}, pages 1925--1934, 2017.

\bibitem[Lin et~al.(2014)Lin, Maire, Belongie, Hays, Perona, Ramanan,
  Doll{\'a}r, and Zitnick]{lin2014microsoft}
Tsung-Yi Lin, Michael Maire, Serge Belongie, James Hays, Pietro Perona, Deva
  Ramanan, Piotr Doll{\'a}r, and C~Lawrence Zitnick.
\newblock Microsoft coco: Common objects in context.
\newblock In \emph{European conference on computer vision}, pages 740--755.
  Springer, 2014.

\bibitem[Long et~al.(2015)Long, Shelhamer, and Darrell]{long2015fully}
Jonathan Long, Evan Shelhamer, and Trevor Darrell.
\newblock Fully convolutional networks for semantic segmentation.
\newblock In \emph{Proceedings of the IEEE conference on computer vision and
  pattern recognition}, pages 3431--3440, 2015.

\bibitem[Mackowiak et~al.(2018)Mackowiak, Lenz, Ghori, Diego, Lange, and
  Rother]{mackowiak2018cereals}
Radek Mackowiak, Philip Lenz, Omair Ghori, Ferran Diego, Oliver Lange, and
  Carsten Rother.
\newblock Cereals-cost-effective region-based active learning for semantic
  segmentation.
\newblock \emph{The British Machine Vision Association BMVC}, 2018.

\bibitem[Mahapatra et~al.(2018)Mahapatra, Bozorgtabar, Thiran, and
  Reyes]{mahapatra2018efficient}
Dwarikanath Mahapatra, Behzad Bozorgtabar, Jean-Philippe Thiran, and Mauricio
  Reyes.
\newblock Efficient active learning for image classification and segmentation
  using a sample selection and conditional generative adversarial network.
\newblock In \emph{International Conference on Medical Image Computing and
  Computer-Assisted Intervention}, pages 580--588. Springer, 2018.

\bibitem[Pardo et~al.(2019)Pardo, Xu, Thabet, Arbelaez, and
  Ghanem]{pardo2019baod}
Alejandro Pardo, Mengmeng Xu, Ali Thabet, Pablo Arbelaez, and Bernard Ghanem.
\newblock Baod: Budget-aware object detection.
\newblock \emph{arXiv preprint arXiv:1904.05443}, 2019.

\bibitem[Ren et~al.(2015)Ren, He, Girshick, and Sun]{ren2015faster}
Shaoqing Ren, Kaiming He, Ross Girshick, and Jian Sun.
\newblock Faster r-cnn: Towards real-time object detection with region proposal
  networks.
\newblock In \emph{Advances in neural information processing systems}, pages
  91--99, 2015.

\bibitem[Ronneberger et~al.(2015)Ronneberger, Fischer, and
  Brox]{ronneberger2015u}
Olaf Ronneberger, Philipp Fischer, and Thomas Brox.
\newblock U-net: Convolutional networks for biomedical image segmentation.
\newblock In \emph{International Conference on Medical image computing and
  computer-assisted intervention}, pages 234--241. Springer, 2015.

\bibitem[Russakovsky et~al.(2015)Russakovsky, Deng, Su, Krause, Satheesh, Ma,
  Huang, Karpathy, Khosla, Bernstein, et~al.]{russakovsky2015imagenet}
Olga Russakovsky, Jia Deng, Hao Su, Jonathan Krause, Sanjeev Satheesh, Sean Ma,
  Zhiheng Huang, Andrej Karpathy, Aditya Khosla, Michael Bernstein, et~al.
\newblock Imagenet large scale visual recognition challenge.
\newblock \emph{International journal of computer vision}, 115\penalty0
  (3):\penalty0 211--252, 2015.

\bibitem[Schwarz et~al.(2018)Schwarz, Milan, Periyasamy, and
  Behnke]{schwarz2018rgb}
Max Schwarz, Anton Milan, Arul~Selvam Periyasamy, and Sven Behnke.
\newblock Rgb-d object detection and semantic segmentation for autonomous
  manipulation in clutter.
\newblock \emph{The International Journal of Robotics Research}, 37\penalty0
  (4-5):\penalty0 437--451, 2018.

\bibitem[Settles(2009)]{settles2009active}
Burr Settles.
\newblock Active learning literature survey.
\newblock Technical report, University of Wisconsin-Madison Department of
  Computer Sciences, 2009.

\bibitem[Shao et~al.(2019)Shao, Wang, and Liu]{shao2019learning}
Jingyu Shao, Qing Wang, and Fangbing Liu.
\newblock Learning to sample: an active learning framework.
\newblock \emph{arXiv preprint arXiv:1909.03585}, 2019.

\bibitem[Siddiqui et~al.(2020)Siddiqui, Valentin, and
  Nie{\ss}ner]{siddiqui2019viewal}
Yawar Siddiqui, Julien Valentin, and Matthias Nie{\ss}ner.
\newblock Viewal: Active learning with viewpoint entropy for semantic
  segmentation.
\newblock In \emph{Proceedings of the IEEE/CVF Conference on Computer Vision
  and Pattern Recognition}, pages 9433--9443, 2020.

\bibitem[Simonyan and Zisserman(2014)]{simonyan2014very}
Karen Simonyan and Andrew Zisserman.
\newblock Very deep convolutional networks for large-scale image recognition.
\newblock \emph{arXiv preprint arXiv:1409.1556}, 2014.

\bibitem[Sinha et~al.(2019)Sinha, Ebrahimi, and Darrell]{sinha2019variational}
Samarth Sinha, Sayna Ebrahimi, and Trevor Darrell.
\newblock Variational adversarial active learning.
\newblock In \emph{Proceedings of the IEEE International Conference on Computer
  Vision}, pages 5972--5981, 2019.

\bibitem[Srivastava et~al.(2014)Srivastava, Hinton, Krizhevsky, Sutskever, and
  Salakhutdinov]{srivastava2014dropout}
Nitish Srivastava, Geoffrey Hinton, Alex Krizhevsky, Ilya Sutskever, and Ruslan
  Salakhutdinov.
\newblock Dropout: a simple way to prevent neural networks from overfitting.
\newblock \emph{The journal of machine learning research}, 15\penalty0
  (1):\penalty0 1929--1958, 2014.

\bibitem[Sugiyama(2006)]{sugiyama2006active}
Masashi Sugiyama.
\newblock Active learning for misspecified models.
\newblock In \emph{Advances in neural information processing systems}, pages
  1305--1312, 2006.

\bibitem[Szegedy et~al.(2015)Szegedy, Liu, Jia, Sermanet, Reed, Anguelov,
  Erhan, Vanhoucke, and Rabinovich]{szegedy2015going}
Christian Szegedy, Wei Liu, Yangqing Jia, Pierre Sermanet, Scott Reed, Dragomir
  Anguelov, Dumitru Erhan, Vincent Vanhoucke, and Andrew Rabinovich.
\newblock Going deeper with convolutions.
\newblock In \emph{Proceedings of the IEEE conference on computer vision and
  pattern recognition}, pages 1--9, 2015.

\bibitem[Vezhnevets et~al.(2012)Vezhnevets, Buhmann, and
  Ferrari]{vezhnevets2012active}
Alexander Vezhnevets, Joachim~M Buhmann, and Vittorio Ferrari.
\newblock Active learning for semantic segmentation with expected change.
\newblock In \emph{2012 IEEE Conference on Computer Vision and Pattern
  Recognition}, pages 3162--3169. IEEE, 2012.

\bibitem[Vijayanarasimhan and Grauman(2009)]{vijayanarasimhan2009s}
Sudheendra Vijayanarasimhan and Kristen Grauman.
\newblock What's it going to cost you?: Predicting effort vs. informativeness
  for multi-label image annotations.
\newblock In \emph{2009 IEEE Conference on Computer Vision and Pattern
  Recognition}, pages 2262--2269. IEEE, 2009.

\bibitem[Worrall et~al.(2017)Worrall, Garbin, Turmukhambetov, and
  Brostow]{worrall2017harmonic}
Daniel~E Worrall, Stephan~J Garbin, Daniyar Turmukhambetov, and Gabriel~J
  Brostow.
\newblock Harmonic networks: Deep translation and rotation equivariance.
\newblock In \emph{Proceedings of the IEEE Conference on Computer Vision and
  Pattern Recognition}, pages 5028--5037, 2017.

\bibitem[Yang et~al.(2017)Yang, Zhang, Chen, Zhang, and
  Chen]{yang2017suggestive}
Lin Yang, Yizhe Zhang, Jianxu Chen, Siyuan Zhang, and Danny~Z Chen.
\newblock Suggestive annotation: A deep active learning framework for
  biomedical image segmentation.
\newblock In \emph{International conference on medical image computing and
  computer-assisted intervention}, pages 399--407. Springer, 2017.

\bibitem[Yoo and Kweon(2019)]{yoo2019learning}
Donggeun Yoo and In~So Kweon.
\newblock Learning loss for active learning.
\newblock In \emph{Proceedings of the IEEE Conference on Computer Vision and
  Pattern Recognition}, pages 93--102, 2019.

\bibitem[Yu et~al.(2019)Yu, Chen, Gkioxari, Bansal, Berg, and
  Batra]{yu2019multi}
Licheng Yu, Xinlei Chen, Georgia Gkioxari, Mohit Bansal, Tamara~L Berg, and
  Dhruv Batra.
\newblock Multi-target embodied question answering.
\newblock In \emph{Proceedings of the IEEE Conference on Computer Vision and
  Pattern Recognition}, pages 6309--6318, 2019.

\bibitem[Zhai et~al.(2019)Zhai, Oliver, Kolesnikov, and Beyer]{zhai2019s4l}
Xiaohua Zhai, Avital Oliver, Alexander Kolesnikov, and Lucas Beyer.
\newblock S4l: Self-supervised semi-supervised learning.
\newblock In \emph{Proceedings of the IEEE international conference on computer
  vision}, pages 1476--1485, 2019.

\bibitem[Zhao et~al.(2017)Zhao, Shi, Qi, Wang, and Jia]{zhao2017pyramid}
Hengshuang Zhao, Jianping Shi, Xiaojuan Qi, Xiaogang Wang, and Jiaya Jia.
\newblock Pyramid scene parsing network.
\newblock In \emph{Proceedings of the IEEE conference on computer vision and
  pattern recognition}, pages 2881--2890, 2017.

\bibitem[Zhou(2012)]{zhou2012ensemble}
Zhi-Hua Zhou.
\newblock \emph{Ensemble methods: foundations and algorithms}.
\newblock CRC press, 2012.

\end{thebibliography}

\newpage
\section*{Supplementary Material}
\section*{Qualitative Results}
\label{S1}
\begin{figure}[h]
\centering
  \includegraphics [scale=.36]{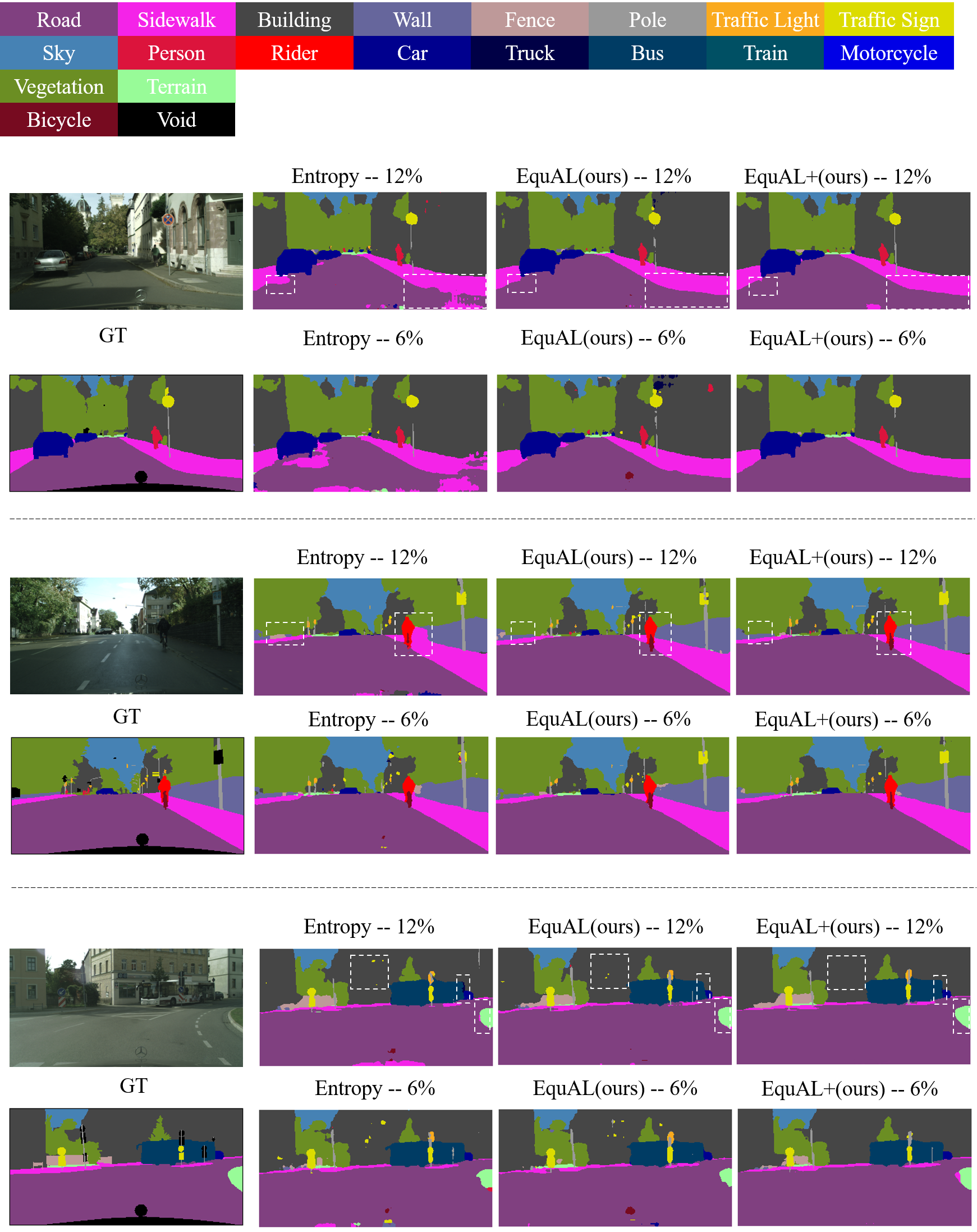}
	\caption{ Qualitative results on  Cityscapes dataset. The second column shows the performance of the entropy baseline. Third and forth columns show the performance of our proposed method.
	For each image the first and second rows show the results for budget of $12\%$  and $6\%$, respectively.
	}
	\label{Sup1}
 \end{figure}

Fig. \ref{Sup1} provides qualitative results for our proposed method comparing to the entropy baseline for two budgets ($6\%$ and $12\%$).
As shown in Fig.  \ref{Sup1}, our method has better qualitative performance for different images.
We provide regions with white dashed lines to highlight the differences among the performances.

 \section*{Numerical Values}
 \label{S2}
 In Fig. 3 in the paper we demonstrate the performance of our proposed method comparing to the other baselines for CamVid and Cityscapes datasets.
 Here, in Table \ref{TB1} and \ref{TB2}, we provide the numerical values corresponding to the Fig 3 in the paper.
 Table \ref{TB1} and \ref{TB2} shows our evaluation for Camvid and Cityscapes, respectively.
 For all the experiments, we report the average and standard deviation of five different runs.

\begin{table}[h]
\centering
\caption{Semantic segmentation performance in terms of mIoU when labeled data is selected using baseline active learning
methods and our method on CamVid dataset.
Bold   entries are
the best and second-best performers.}
\resizebox{4 in}{!} {
\begin{tabular}{ccccccc}
\hline 
\multicolumn{7}{c}{CamVid}\tabularnewline
\hline 
\hline 
\multirow{2}{*}{Models} & \multicolumn{6}{c}{Budget}\tabularnewline
\cline{2-7} \cline{3-7} \cline{4-7} \cline{5-7} \cline{6-7} \cline{7-7} 
 & 8\% & 12\% & 16\% & 20\% & 24\% & 100\%\tabularnewline
\hline 
Fully-Supervised & - & - & - & - & - & 0.6450$\pm$0.017\tabularnewline
Fully-Supervised+ & - & - & - & - & - & 0.6724$\pm$0.006\tabularnewline
\hline 
Random & 0.5189$\pm$0.0265 & 0.5322$\pm$0.0282 & 0.5501$\pm$0.0248 & 0.5821$\pm$0.0135 & 0.5911$\pm$0.0244 & -\tabularnewline
Entropy & 0.5451$\pm$0.0176 & 0.6111$\pm$0.01167 & 0.6192$\pm$0.0189 & 0.6311$\pm$0.0083 & 0.6347$\pm$0.0089 & -\tabularnewline
EquAL (ours) & \textbf{0.5673$\pm$0.0112} & \textbf{0.6182$\pm$0.0062} & \textbf{0.6274$\pm$0.0148} & \textbf{0.6341$\pm$0.0038} & \textbf{0.6385$\pm$0.0064} & -\tabularnewline
EquAL+ (ours) & \textbf{0.6213$\pm$0.0047} & \textbf{0.6446$\pm$0.0038} & \textbf{0.6492$\pm$0.0065} & \textbf{0.6560$\pm$0.0048} & \textbf{0.6630$\pm$0.0081} & -\tabularnewline
\hline 
\end{tabular}
}
\label{TB1}
\end{table}

\begin{table}[h]
\centering
\caption{Semantic segmentation performance in terms of mIoU when labeled data is selected using baseline active learning
methods and our method on CamVid dataset.
Bold   entries are
the best and second-best performers.}
\resizebox{5 in}{!} {

\begin{tabular}{ccccccccc}
\hline 
\multicolumn{9}{c}{Cityscapes}\tabularnewline
\hline 
\multirow{2}{*}{Model} & \multicolumn{8}{c}{Budget}\tabularnewline
\cline{2-9} \cline{3-9} \cline{4-9} \cline{5-9} \cline{6-9} \cline{7-9} \cline{8-9} \cline{9-9} 
 & 1\% & 2\% & 3\% & 6\% & 9\% & 12\% & 15\% & 100\%\tabularnewline
\hline 
Fully-Supervised & - & - & - & - & - & - & - & 0.6346$\pm$0.0031\tabularnewline
Fully-Supervised+ & - & - & - & - & - & - & - & 0.6687$\pm$0.0057\tabularnewline
\hline 
Random & 0.3346$\pm$0.0248 & 0.426$3\pm$0.0233 & 0.4661$\pm$0.0140 & 0.5202$\pm$0.0211 & 0.5372$\pm$0.0152 & 0.5518 $\pm$0.0134 & 0.5611$\pm$0.0082 & -\tabularnewline
Entropy & 0.3903$\pm$0.0066 & 0.4750$\pm$0.0034 & 0.5072$\pm$0.0042 & 0.5671$\pm$0.0022 & 0.5801$\pm$0.0038 & 0.6003$\pm$0.0051 & 0.6151$\pm$0.0060 & -\tabularnewline
EquAL (ours) & \textbf{0.4242$\pm$0.0033} & \textbf{0.4880$\pm$0.0069} & \textbf{0.5184$\pm$0.0045} & \textbf{0.5713$\pm$0.0049} & 0.5890$\pm$0.0046 & \textbf{0.6065$\pm$0.0025} & \textbf{0.6185$\pm$0.0062} & -\tabularnewline
EquAL+ (ours) & \textbf{0.4885$\pm$0.0035} & \textbf{0.5590$\pm$0.0030} & \textbf{0.5809$\pm$0.0060} & \textbf{0.6224$\pm$0.0062} & \textbf{0.6352$\pm$0.0053} & \textbf{0.6446$\pm$0.0034} & \textbf{0.6562$\pm$0.0014} & -\tabularnewline
\hline 
\end{tabular}

}
\label{TB2}
\end{table}

 In both of the tables Fully-Supervised denotes the maximum achievable network
performance when all the data in the training set are labeled. 
Fully-Supervised baseline uses cross-entropy loss during the training.
 Furthermore, Fully-Supervised+ denotes the   maximum achievable network
performance when all the data in the training set are labeled and cross-entropy loss as well as pixel-wise self-consistency loss is used during the training.

 After active sample selection is finished and the budget is met, the final labeled dataset has images that are mostly partially labeled. 
 The CE loss uses only the labeled  part of the images during  training.
 In our proposed method we claim that it is important to use the unlabeled regions as well. 
 Therefore, we further proposed EquAL+, where it uses both CE and pixel-wise self-consistency during the final retraining stage.
 The pixel-wise self-consistency loss is independent of the labeled ground truth and regularizes the network based on the predictions between the results from the input image and its horizontally-flipped transformed version (Eq 5c  in the paper).

 



\end{document}